\documentclass[10pt,twocolumn,letterpaper]{article}

\usepackage{wacv}              

\usepackage{multirow}
\usepackage[table]{xcolor}
\usepackage{amssymb}
\usepackage{comment}
\usepackage[numbers,sort&compress]{natbib}
\definecolor{wacvblue}{rgb}{0.21,0.49,0.74}
\usepackage[pagebackref,breaklinks,colorlinks,allcolors=wacvblue]{hyperref}

\DeclareRobustCommand*{\IEEEauthorrefmark}[1]{%
  \raisebox{0pt}[0pt][0pt]{\textsuperscript{\footnotesize #1}}%
}

\def\wacvPaperID{960} 
\def\confName{WACV}
\def\confYear{2026}

\title{Visual Detector Compression via Location-Aware Discriminant Analysis}

\renewcommand{\thefootnote}{\fnsymbol{footnote}} 

\author{
    Qizhen Lan\IEEEauthorrefmark{1,2}\thanks{The two student authors contributed equally.},
    Jung Im Choi\IEEEauthorrefmark{2}\textsuperscript{\thefootnote}, 
    \setcounter{footnote}{1} 
    Qing Tian\IEEEauthorrefmark{1,2}\thanks{Corresponding author.}
  \\
  \IEEEauthorrefmark{1}Dept. of Computer Science, University of Alabama at Birmingham, Alabama, USA\\
  \IEEEauthorrefmark{2}Dept. of Computer Science, Bowling Green State University, Ohio, USA\\
  \tt\small{qlan@uab.edu, choij@bgsu.edu, qtian@uab.edu}
}
\renewcommand{\thefootnote}{\arabic{footnote}}

\begin{document}
\maketitle
\begin{abstract}
Deep neural networks are powerful, yet their high complexity greatly limits their potential to be deployed on billions of resource-constrained edge devices. Pruning is a crucial network compression technique, yet most existing methods focus on classification models, with limited attention to detection. Even among those addressing detection, there is a lack of utilization of essential localization information. Also, many pruning methods passively rely on pre-trained models, in which useful and useless components are intertwined, making it difficult to remove the latter without harming the former at the neuron/filter level. To address the above issues, in this paper, we propose a proactive detection-discriminants-based network compression approach for deep visual detectors, which alternates between two steps: (1) maximizing and compressing detection-related discriminants and aligning them with a subset of neurons/filters immediately before the detection head, and (2) tracing the detection-related discriminating power across the layers and discarding features of lower importance. Object location information is exploited in both steps. Extensive experiments, employing four advanced detection models and four state-of-the-art competing methods on the KITTI and COCO datasets, highlight the superiority of our approach. Remarkably, our compressed models can even beat the original base models with a substantial reduction in complexity.
\end{abstract}    
\section{Introduction}
\label{sec:intro}

Visual detection is crucial to many real-world applications and usually, it needs to be performed both accurately and in real time. For example, when it comes to self-driving perception, even a few milliseconds’ delay can make the difference between life and death. Visual detection is a more challenging task than visual classification because classification only needs to predict a label for one target (i.e., the whole image) while detection needs to produce both a label and a bounding box for each possible target. The number of possible targets is orders of magnitudes larger in detection than in classification, i.e., $O(p^2)$ vs. $O(1)$ where $p$ is the number of image pixels.
Deep neural networks have revolutionized the field of computer vision with their high accuracy, but most are expensive in terms of size, computation, and energy consumption and are thus impractical for devices without high-end GPUs. To address this issue, many pruning approaches have been proposed to discard ``unimportant'' parameters or filters.

\begin{figure}[t]
\begin{center}
\includegraphics[width=\linewidth]{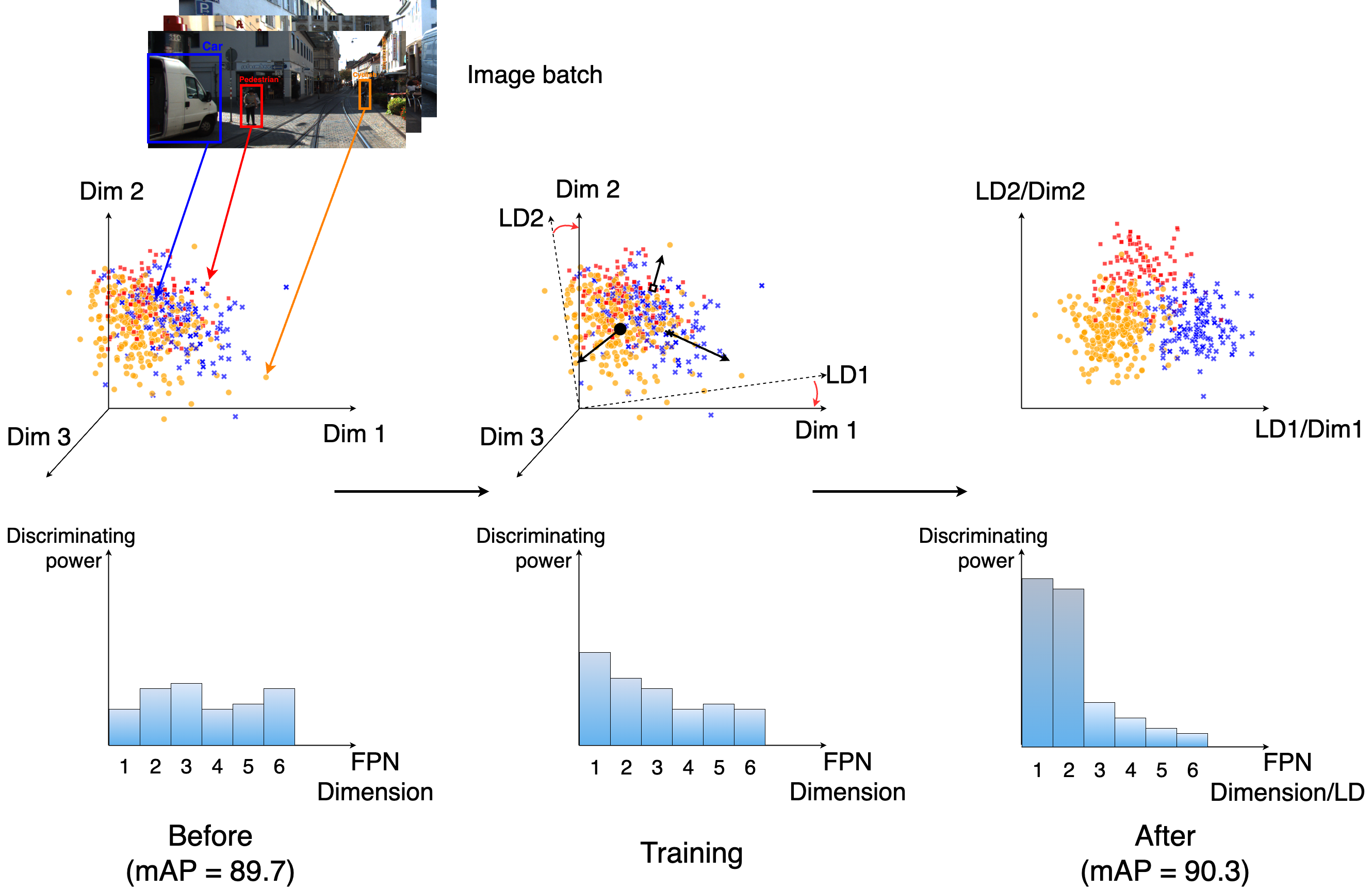}
\end{center}
\caption{Effects of our Location-aware Discriminant Training (LDT) on class separation, discriminant optimization, and detection performance. This experiment is conducted on KITTI using the GFL \cite{li2020generalized} base detector. The LDT loss is calculated at the neck layers of each scale and exerts influence over the layers during (re)training. It maximizes and condenses detection-related discriminants, and aligns them with filter/channel dimensions for safe structured pruning. By reducing the number of required dimensions, it makes more room for pruning (details in Sec. \ref{ssec:ldt}). Also, the filters to be retained are more refined. Non-discriminative dimensions approach zero and are omitted for clarity.}
\label{fig:LDT KITTI}
\end{figure}
However, importance measures for pruning are often computed locally (with relationships within/across filters/layers ignored) and some do not necessarily reflect the final task utility \cite{hu2016network, he2019filter, lin2020hrank, iurada2024finding}. Also, most of the existing pruning approaches depend on a pre-trained or passively learned model. It may be too late to prune after the fact that useful and harmful components are already intertwined, e.g., at the filter level. Most existing structured pruning approaches assume that a filter is either entirely useful or completely useless so that pruning or retaining the entire filter is well justified. As we will show later in Fig. \ref{fig:LDT KITTI}, filter utility is not binary.
Furthermore, almost all existing pruning approaches focus on the relatively simple classification task. Among the few pruning works that tackle object detection, most directly employ the importance measures for classification tasks with no or few modifications \cite{o2018evaluating, yao2021joint}. They do not explicitly consider location information, even if it is essential to the task of object detection.

To address these limitations, we propose location-aware discriminant analysis for deep detector compression, which actively concentrates and refines detection-relevant discriminants into a compact set of latent channel dimensions before treating detector pruning as a structured dimensionality reduction problem within the deep feature space.

Visual detection involves two closely related subtasks, i.e., localization and classification. Accurate localization depends on correctly identifying discriminative features for each class, while accurate classification depends on the precise localization of these features. We quantify the visual detection utility (i.e., the class separation and localization power) using location-aware deep discriminant analysis and maximize the utility during training. In addition, to help compress utility and reduce redundancy, we also introduce a correlation penalty term to the training objective. The two objective terms will synergistically align the utility with a condensed subset of latent filters/channels (Sec. \ref{ssec:ldt}), simultaneously enabling and creating more room for structured pruning.
Specifically, we ``push'' the detection discriminative power into alignment with a subset of neck output channels immediately preceding the detection head, making dimension selection at the filter/channel level safe. Given that there are theoretically at most \#classes - 1 discriminants, our training with the covariance penalty term suppresses the activity of non-discriminating dimensions. Through a real example, Fig.~\ref{fig:LDT KITTI} illustrates the impact of our Location-aware Discriminant Training (LDT) in boosting and condensing detection-related discriminating power, while also aligning discriminants with filters/channels for structured detector pruning.
Our LDT procedure ensures that filters to be preserved are more refined while those to be pruned have minimal utility, thereby better justifying the discarding/retaining of an entire filter.
 
After the detection-related discriminants are reorganized through our LDT, pruning begins with discarding filter/channel dimensions with low discriminating power in the neck (e.g., FPN \cite{fpn}) output. With the help of derivative-based dependency and object location information, we then trace the detection-related discriminating power back to the preceding layers' filters to guide the pruning process.
The main contributions of this paper are as follows: 
\begin{itemize}
\item We propose a novel model compression approach for deep visual detectors based on deep location-aware discriminant analysis. Unlike conventional structured pruning methods, we proactively maximize the room for detector pruning through active discriminant maximization, compression, and alignment during (re)training. Our method offers a better justification for performing pruning at the filter level.
\item Our method is a pioneering pruning technique that is specifically designed for visual detectors and explicitly utilizes location information. We propose to integrate object location information into both processes: deep location-aware discriminant training and channel importance calculation. We find that incorporating location information consistently improves the performance of our compressed models.
\item Extensive experiments on the KITTI and COCO datasets, using four base detectors and four competing state-of-the-art methods, demonstrate the efficacy of our proposed approach. Some of our pruned models even outperform the original models despite having much less complexity. For instance, on the KITTI dataset, one of our pruned GFL models with 70.0\% parameters removed outperforms the unpruned model by 0.7\% mAP. A pruned YOLOX-S model beats its original base model by 1.6\% mAP, with a 34.8\% reduction in model size.
\end{itemize}

\section{Related Work}
\label{sec:re_work}

Pruning aims to remove as many redundant or unimportant units as possible while maintaining the original model's performance. There are two widely discussed types of pruning, i.e., unstructured pruning and structured pruning. Inspired by early works that remove weights individually~\cite{LeCun1989, Hassibi1992}, various unstructured pruning methods have been proposed~\cite{chen2015compressing,lee2021layer, yang2017designing, xu2023efficient, Zhang2018, gao2023structural}. For example, Gao \etal \cite{gao2023structural} presented a jointly optimized, layer-adaptive weight pruning approach to minimize output distortion while pruning weights across all layers. However, these methods introduce irregular sparsity, requiring specialized software and/or hardware support~\cite{Park2017, han2016eie}, which in turn adds computational overhead.

Unlike unstructured methods, structured pruning techniques remove entire filters or neurons in a network~\cite{Li2017, Tian2017deep, Xu2023, lin2020hrank, Sui2021,tian2023grow}. Such methods lead to direct computational and storage reductions on general hardware, making them more practical for real-world deployment. The key difference among existing structured pruning methods lies in the way to quantify the importance of a filter/channel, such as using L1-norm of filters~\cite{Li2017}, sparsity of output feature maps~\cite{hu2016network}, parameter sparsity~\cite{Wen2016}, and loss drop w.r.t. filter removal~\cite{luo2017thinet}. Lin \etal \cite{lin2020hrank} introduced HRank, which assumes that low-rank feature maps carry less information.
Liu \etal \cite{liu2021group} presented Group Fisher Pruning to identify and prune coupled channels, leveraging Fisher information. Sui \etal \cite{Sui2021} measured the correlations among feature maps, considering those with high correlation to others as less informative. Xu \etal \cite{Xu2023} proposed a partial regularization technique to improve performance after one-shot pruning. Iurada \etal \cite{iurada2024finding} proposed Path eXclusion (PX) algorithm using Neural Tangent Kernel theory~\cite{jacot2018ntk} to align sparse and dense network training dynamics. These methods often rely on locally computed importance measures that are not directly related to the final task utility.
Moreover, most current pruning methods have focused on classification tasks, paying no or limited attention to object detection. Among the few works handling object detection tasks, Xie \etal \cite{Xie2020} failed to consider the joint effect of multiple layers, and Choi and Tian \cite{Choi2023} took a passive strategy and compressed all layers equally. 
More importantly, although neuron/filter-based pruning can lead to direct savings on general machines, most, if not all, such approaches ignore the fact that most filters are not entirely useful or completely useless. It is inevitable to throw away some useful components when pruning an entire neuron. In this paper, targeting visual object detection, we propose to first maximize and condense the location-aware discriminants and align them with a compressed set of channel dimensions across different scales before detection-discriminants-based pruning can be conducted at the filter/channel level in a safer manner.
\begin{figure*}[t]
\begin{center}
\includegraphics[width=\linewidth,trim={0.55in 0.1in 0.4in 0.085in},clip]{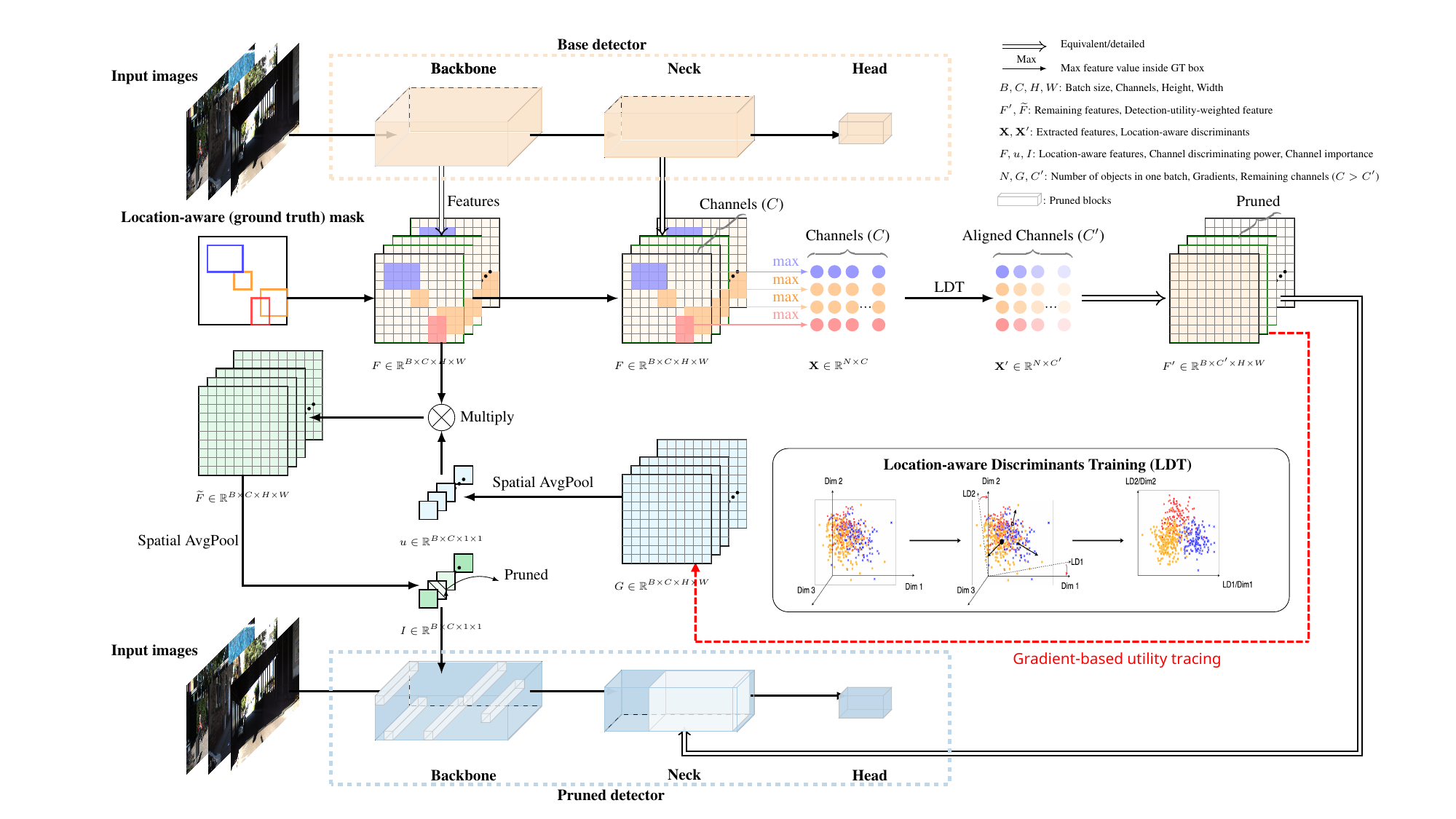}
\end{center}
\caption{Overview of the proposed location-aware discriminant training and pruning framework. Different colors of bboxes/dots represent different object categories. After LDT, the color intensity serves as an indicator of a channel's detection-related discriminating power. Channels with darker colors possess greater discriminative power for detection.}
\label{fig:overview}
\end{figure*}
\section{Methodology}
\label{sec:method}

Precise localization of a target class relies on correctly identifying its discriminative features \cite{xie2017adversarial}. 
In this paper, we treat pruning as a discriminants reorganization and reduction problem. 
In Sec. \ref{ssec:ldt}, we propose location-aware deep discriminant analysis to maximize the class separation and localization power on each feature scale during training. 
This step also better organizes the information flow transferred across the network and makes more room for structured pruning. 
In Sec. \ref{ssec:tracingpruning}, we trace detection-related discriminants across the network before pruning the less important filters/channels.
Fig.~\ref{fig:overview} offers an overview of our proposed method.

\subsection{Location-aware Discriminant Training}\label{ssec:ldt}

Our Location-aware Discriminant Training (LDT) optimizes useful feature representations at the neck (e.g., FPN) output, positioned directly before the detection heads, while extending its influence across the network layers. The training process guided by our designed loss terms gradually maximizes the separation between distinct object categories' features and minimizes their redundancy. In the meanwhile, LDT also contributes to the alignment of detection-related discriminants and a compressed set of latent channels.

To perform deep location-aware discriminant analysis on object detectors, we first utilize the bounding box information to extract relevant object features. For each neck scale's output, we employ a mask per object, denoted as $\mathbf{M}_{i,j,n}$, to identify the ground truth bounding box positions in the image:

\begin{equation}
    \mathbf{M}_{i,j,n}= \begin{cases}
    1,\quad & \text{if}\quad (i,j)\in \mathcal{G}_n\\
    0,\quad & \text{otherwise}
\end{cases} \,.
\end{equation}

\noindent This mask takes a value of 1 when the position $(i, j)$ falls within the ground truth bounding box $\mathcal{G}_n$, and 0 otherwise. $n$ indicates the $n$-th object in one training batch.
We can use such masks to create location-aware features for specific objects:

\begin{equation}
\mathbf{A}_{i,j,n, \theta} = \mathbf{M}_{i,j,n} F_{i,j, \theta}\,.
\end{equation}

\noindent where $F_{i,j, \theta}$ represents the feature (vector of dimension $C$) at position $(i, j)$ given the model parameter setting $\theta$. For each object instance $n$, we obtain a 1-D feature vector $\mathbf{v}_n$ by taking the max value of each channel within the corresponding bounding box region:
\begin{equation}
\mathbf{v}_{n, \theta} = \underset{ i,j}{\text{max}}(A_{i, j, n, \theta})\,.
\end{equation}

\noindent We then stack such vectors of $N$ objects into a matrix $\mathbf{X}_{\theta} = \left[\mathbf{v}_{1, \theta}, \mathbf{v}_{2, \theta}, \ldots, \mathbf{v}_{N, \theta}\right]^{T}$, which, of dimension $(N, C)$, serves as the input for later steps.

For object detection, we propose to capture and maximize the class separation and localization power using location-aware deep discriminant analysis. Through optimizing detector parameters $\theta$, our goal is to maximize:

\begin{equation}\label{eq:ratio}
\mathbf{J}(\theta) = |\frac{\mathbf{W}_{\theta}^{T}\mathbf{S}_{b,\theta}\mathbf{W}_{\theta}}{\mathbf{W}_{\theta}^{T}\mathbf{S}_{w,\theta}\mathbf{W}_{\theta}}|\,,
\end{equation}

\noindent where $\mathbf{S}_{w,\theta}$ and $\mathbf{S}_{b,\theta}$ are within-class and between-class scatter matrices\footnote{Here, the assumption-based versions of the scatter matrices from \cite{dorfer2015deep} are adopted.}:
\begin{equation}
    \mathbf{S}_{w, \theta} = \frac{1}{G} \sum_{g=1}^{G} \Bar{\mathbf{X}}_{g, \theta}^T\Bar{\mathbf{X}}_{g, \theta}\,,
\end{equation}

\begin{equation}\label{eq: Sb}
    \mathbf{S}_{b, \theta} = \mathbf{S}_{t, \theta} - \mathbf{S}_{w, \theta}\,,
\end{equation}

\noindent where $G$ denotes the number of classes in a batch and $\mathbf{S}_{t, \theta}$ represents the total scatter matrix:

\begin{equation}
    \mathbf{S}_{t, \theta} = \frac{1}{N-1} {\Bar{\mathbf{X}}_{\theta}}^T {\Bar{\mathbf{X}}_{\theta}}\,,
\end{equation}

\noindent where $\Bar{\mathbf{X}}_{\theta}$ indicates the mean-centered feature matrix given the parameter setting $\theta$. The subscript $g$, if any, indicates the class. For data $X$, the centering operation is defined as:

\begin{equation}
    \mathbf{\Bar{X}} = (\mathbf{I}_N - N^{-1}1_N 1_{N}^{T})\mathbf{X}\,,
\end{equation}

\noindent where $N$ is the number of observations in X, $1_N$ stands for a $N\times1$ vector of ones.

Our objective is to maximize the location-aware discriminating power (Eq. \ref{eq:ratio}) via training, which comes down to solving the following generalized eigenvalue problem:

\begin{equation}\label{eq:lda}
\mathbf{S}_{b,\theta}v_{\theta} = \lambda_{\theta} \mathbf{S}_{w,\theta}v_{\theta}\,,
\end{equation}

\noindent where the eigenvector $v_{\theta}$ is a $\mathbf{W}_{\theta}$ column. The corresponding eigenvalue $\lambda_\theta$ indicates the location-aware discriminating power along the eigenvector direction. We achieve our objective by maximizing the top $K$ eigenvalues $\lambda_{k, s, \theta}$, emphasizing those dimensions that contribute most significantly to the location-aware separability of object categories. Typically, modern detectors have a multi-scale neck, and we perform deep location-aware discriminant analysis at each scale. We define our Location-aware Discriminating (LD) loss $\mathcal{L}_{LD}$ as:

\begin{equation} \label{eq:ldt}
\mathcal{L}_{LD} = -\frac{1}{K} \sum_{s=1}^{S} \sum_{k=1}^{K} \lambda_{k, s, \theta}\,,
\end{equation}

\noindent where $s$ denotes a scale in the detector's neck. The threshold for choosing the top $K$ eigenvalues for scale $s$ is determined by $\phi \cdot \max(\lambda_{k, s,\theta})$, where $\phi$ is a hyperparameter. 
To reduce redundancy among different dimensions, we also introduce a covariance penalty term $\mathcal{L}_{cov}$, which is defined as:

\begin{equation} \label{eq:cov}
\mathcal{L}_{cov} = \sum_{s=1}^{S} \left\| \mathbf{S}_{t,s,\theta} - \text{diag}(\mathbf{S}_{t,s, \theta}) \right\|_1\,,
\end{equation}

\noindent where $\text{diag}(\mathbf{S}_{t, s})$ extracts the diagonal elements of the total scatter matrix $\mathbf{S}_{t, s}$.
It is worth noting that the two terms ($\mathcal{L}_{LD}$ and $\mathcal{L}_{cov}$) have a synergistic effect. Training with them jointly offers two benefits: (1) the two terms maximize and condense the location-aware discriminating power, making more room for pruning. They also introduce regularization into the model and help put non-discriminant dimensions dormant. (2) the two terms align the discriminants with a compressed set of neck output channels, which disentangles the utility and makes pruning at the filter/channel level safe.
To be more specific, $\mathcal{L}_{cov}$ helps de-correlate filter/channel dimensions in the neck output. In Eq. \ref{eq:lda}, if the dimensions in the two scatter matrices are de-correlated, the two matrices are diagonal, and the resulting eigenvectors, or $\mathbf{W}_{\theta}$ columns, are standard basis vectors (i.e., the discriminant-channel alignment is realized). In addition to better preparing the model for structured pruning, including $\mathcal{L}_{LD}$ and $\mathcal{L}_{cov}$ in the training objective leads to consistent improvement over the base, as will be demonstrated across our experiments in Sec.~\ref{sec:experiments}.
To sum up, the total loss of our LDT process is as follows:

\begin{equation} \label{eq:total}
\mathcal{L}_{LDT} = \mathcal{L}_{det} + \alpha\mathcal{L}_{LD} + \beta\mathcal{L}_{cov}\,,
\end{equation}

\noindent where $\mathcal{L}_{det}$ denotes the standard detection loss and $\alpha$ and $\beta$ are balancing hyper-parameters. Our LDT lays the foundation for cross-layer detection-related discriminant tracing and pruning at the neuron/filter level.


\subsection{Detection-related Discriminant Tracing and Pruning}\label{ssec:tracingpruning}

After our LDT process on a visual detector, the alignment of detection-related discriminants and a subset of neck output channels is realized at each scale. 
Theoretically, there are at most \#classes - 1 discriminants. In conjunction with the LD term, the correlation penalty term helps suppress the activation of dimensions that lack discriminative power. This facilitates the pruning of channels with lower detection-related discriminatory power at the top of the neck. To trace the discriminating power to contributing layers, we leverage first-order derivatives. We compute the contribution of the $c$-th channel in layer $l$ to the detection-related discriminating power as:
\begin{equation} \label{eq: utility_weights}
    u^{(l)}_{c} = \sum_s\frac{1}{H \cdot W} \sum_{i=1}^{H} \sum_{j=1}^{W} \frac{\partial F'_{neck,s}}{\partial {F}^{(l)}_{c,i,j}}\,,
\end{equation}

\noindent where $F'_{neck,s}$ denotes the post-LDT discriminative features produced by the detector’s neck at scale $s$. $F^{(l)}_c$ represents the $c$-th feature map of layer $l$, and $W$ and $H$ stand for the width and height of $F^{(l)}_c$, respectively. The channel's detection utility is derived from the neck output of each relevant scale and summed up over all such scales. To highlight the features that contribute more to the detection-related discriminating power, we define the detection-utility-weighted feature map as:
\begin{equation} \label{eq: utility_map}
    \widetilde{F^{(l)}_{c,i,j}} = u^{(l)}_c F^{(l)}_{c,i,j}\,.
\end{equation}

\begin{figure*}[htp!]
\begin{center}
   \begin{subfigure}{0.33\textwidth}
     \includegraphics[width=\linewidth]{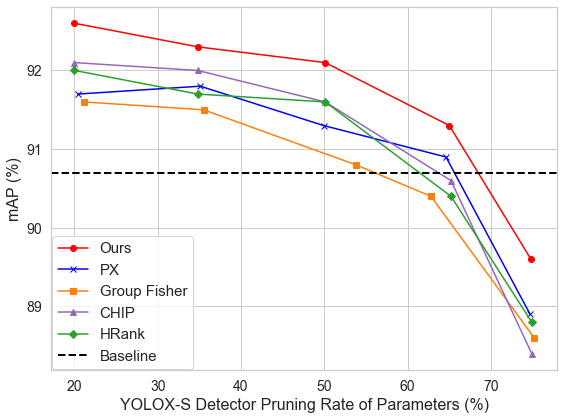}
     \caption{YOLOX-S} \label{fig:yolox}
   \end{subfigure}  
  \begin{subfigure}{0.33\textwidth}
     \includegraphics[width=\linewidth]{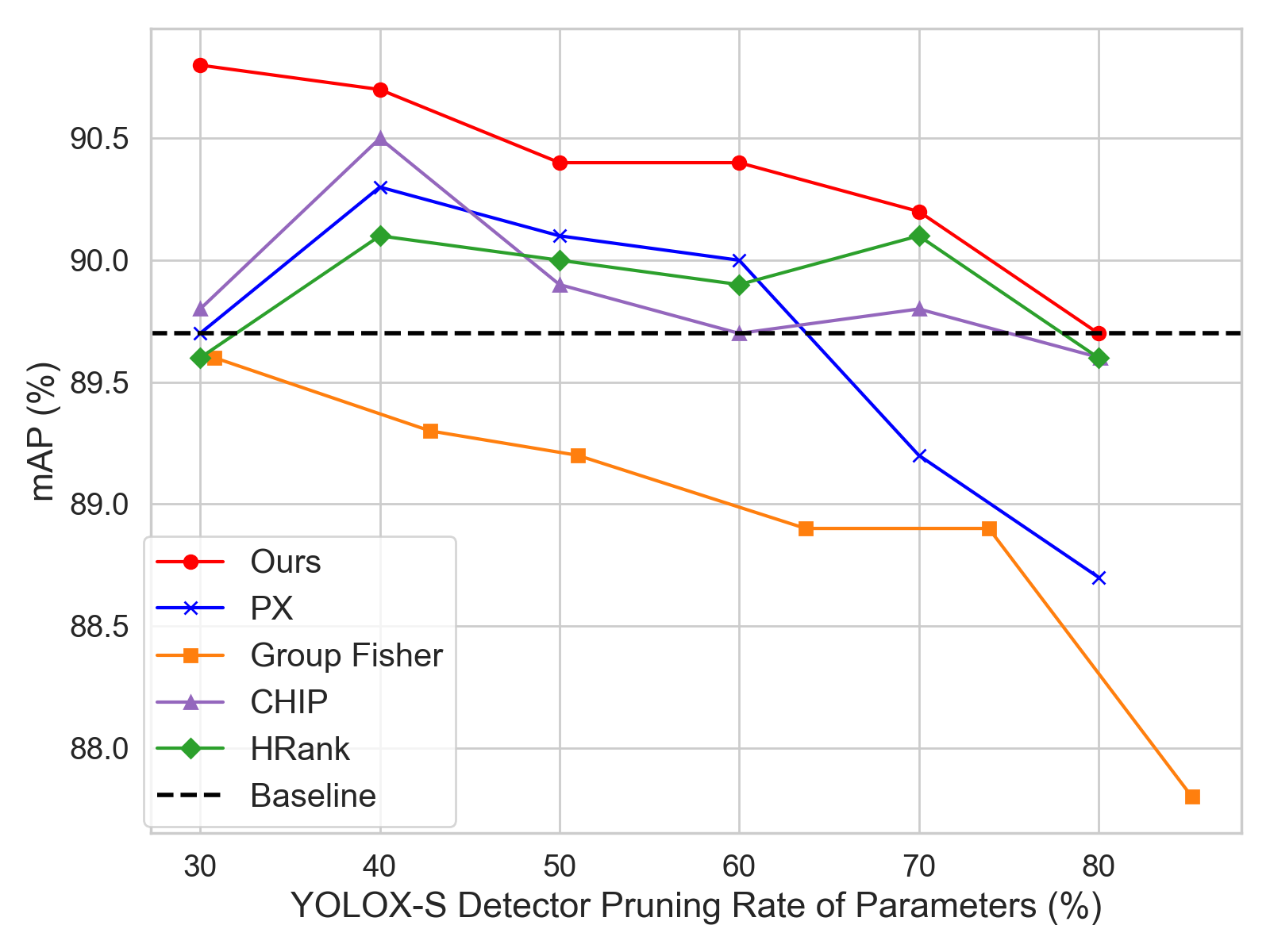}
     \caption{ATSS} \label{fig:atss}
   \end{subfigure}
   \begin{subfigure}{0.33\textwidth}
     \includegraphics[width=\linewidth]{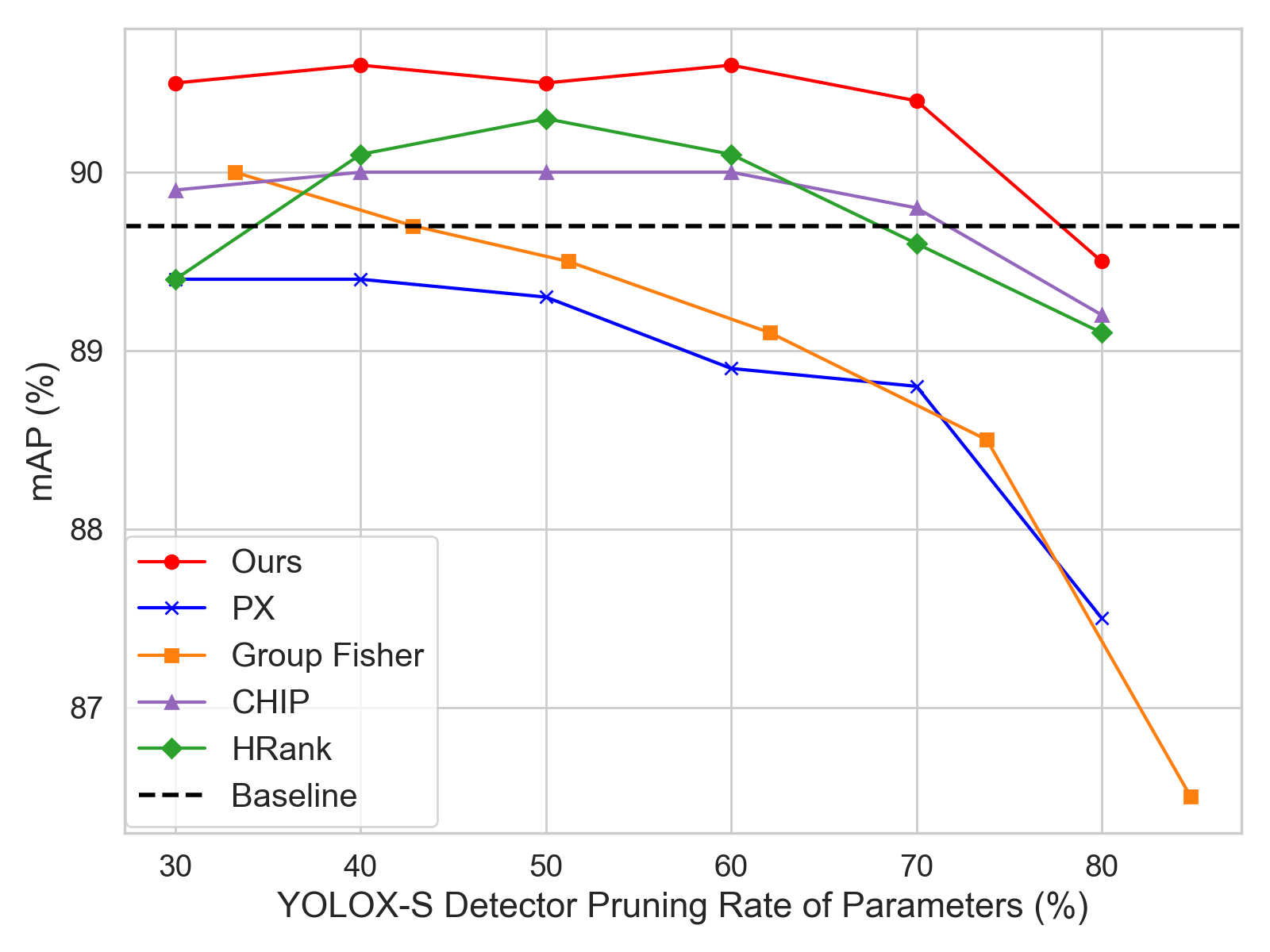}
     \caption{GFL} \label{fig:gfl}
   \end{subfigure}
\end{center}
\caption{
mAP vs. parameter pruning rate on the KITTI dataset, evaluated using the (a) YOLOX-S, (b) ATSS, and (c) GFL detectors. The black dashed line represents the unpruned base models' mAPs: 90.7\% for YOLOX-S and 89.7\% for ATSS and GFL. Our pruning method outperforms the four competing methods, i.e., HRank~\cite{lin2020hrank},
GroupFisher~\cite{liu2021group}, CHIP~\cite{Sui2021}, and PX~\cite{iurada2024finding}, across different base detectors and various pruning rates. All pruning methods are applied iteratively.} \label{fig:kitti}
\end{figure*}

In addition, to minimize irrelevant distractions, we place greater emphasis on object locations and surrounding context when calculating channel importance.
The attention allocated to position $(i, j)$ due to the $n$-th object is defined as:
\begin{equation} \label{eq: location_mask}
    \mathcal{M}^{(l)}_{i,j,n}=
    \begin{cases}
    1,\quad & \text{if}\, (i, j)\in \mathcal{G}_n\\
    a \big((i-x_n)^2+(j-y_n)^2\big)^{-b},\quad & \text{otherwise}
    \end{cases}
\end{equation}

\noindent where $(x_n, y_n)$ is the center point of $n$-th ground truth bounding box $\mathcal{G}_n$. The attention degrades as the pixel moves away from the bounding box. Hyperparameters $a$ and $b$ control the attention decay rate. If a pixel receives attention from multiple bounding boxes, we only consider the bounding box producing the highest attention. By integrating location information into the importance measure, we direct more attention to the foreground objects than their periphery.

We establish the channel importance of the $c$-th channel in layer $l$ as follows:

\begin{equation} \label{eq: importance}
    I^{(l)}_c = \frac{1}{D}\sum_{d}\sum_{i}\sum_{j}\mathcal{M}^{(l)}_{i,j,d} \widetilde{F^{(l)}_{c,i,j,d}}\,,
\end{equation}

\noindent where index $d$ refers to the $d$-th image among a set of $D$ input images. In Sec. \ref{subsec:ablation}, we will demonstrate that a small number of input images are enough to estimate the channel's detection-related utility. 
Based on our channel importance score, we discard less informative channels and their corresponding filters.
It is worth noting that the overhead of LDT training and discriminant tracing is negligible and incurred only during offline derivation of the detector. The resulting compressed detector incurs no runtime overhead.
\section{Experiments and Results}
\label{sec:experiments}

\subsection{Experimental Setup}

\paragraph{Datasets and base models}

To demonstrate the efficacy of our proposed detector compression method, we adopt two popular visual object detection datasets, i.e., KITTI~\cite{lin2014microsoft} and Microsoft COCO~\cite{lin2014microsoft}. For the KITTI dataset, we follow the category grouping convention of \cite{lan2022instance} and split it into training and validation sets with an 8:2 ratio. 
On the two datasets, we compare the performance of our method with four state-of-the-art approaches (i.e., HRank~\cite{lin2020hrank}, GroupFisher~\cite{liu2021group}, CHIP~\cite{Sui2021}, and PX~\cite{iurada2024finding}), using four different base detectors (i.e., GFL \cite{li2020generalized}, ATSS\cite{zhang2020bridging}, YOLOX \cite{ge2021yolox}, and YOLOF \cite{chen2021you}). These base models were chosen for their diverse architectures and effectiveness in real-time, resource-limited environments.

\paragraph{Implementation details}

Our implementations are based on the MMDetection framework \cite{mmdetection}. 
We set $\phi$ = 5e-3, $a$ = 1.2, $b$ = 0.062, $\alpha$ = 5e-4, and $\beta$ = 7.5 for all experiments. 
All models are retrained after compression to restore their performance. 
For our method, the LDT is applied to both training and re-training.
During LDT, YOLOX-S is trained for 100 epochs with SGD (lr = 0.01, momentum = 0.9, weight decay = 5e-4, batch size = 16), while YOLOF, GFL, and ATSS use the same settings but for 24 epochs with weight decay of 1e-4.




\subsection{Quantitative Results}

\subsubsection{Results on KITTI}

As shown in Fig.~\ref{fig:kitti}, on KITTI, our pruning method outperforms all the four competing methods across different base detectors and various parameter pruning rates. In addition, some of our pruned models surpass the performance of the unpruned base detectors, represented by the dashed line. For example, as illustrated in Fig.~\ref{fig:kitti} a, our pruned YOLOX-S detector, when pruned at a rate of 65\%, attains a 0.6\% increase in mAP. Similarly, our pruned ATSS (Fig.~\ref{fig:kitti} b) and GFL (Fig.~\ref{fig:kitti} c) detectors, with a 70\% pruning rate, achieve mAP improvements of 0.7\% and 0.5\%, respectively. 

\subsubsection{Results on COCO}

\begin{figure*}[!htp]
\begin{center}
   \begin{subfigure}{0.24\textwidth}
     \includegraphics[width=\linewidth, height=.9in]{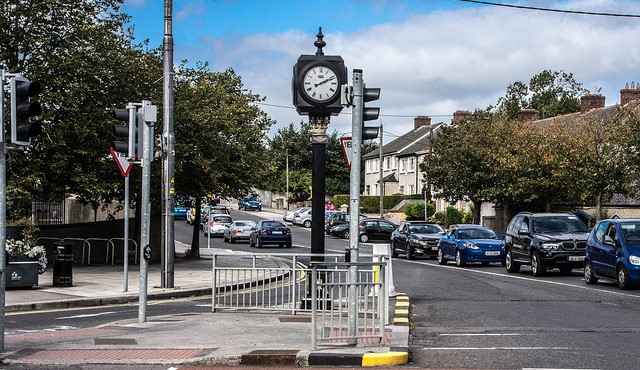}
   \end{subfigure}
   \begin{subfigure}{0.24\textwidth}
     \includegraphics[width=\linewidth, height=.9in]{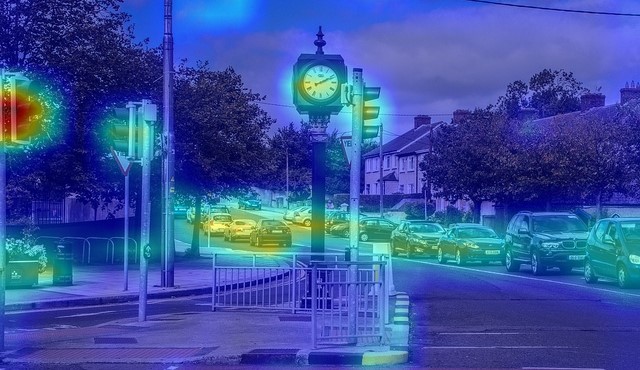}
   \end{subfigure}
   \begin{subfigure}{0.24\textwidth}
     \includegraphics[width=\linewidth, height=.9in]{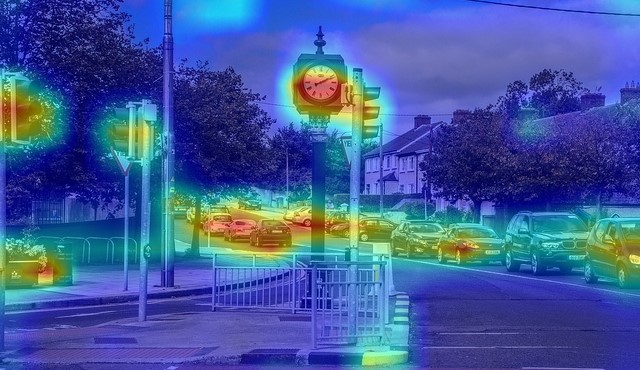}
   \end{subfigure}
   \begin{subfigure}{0.24\textwidth}
     \includegraphics[width=\linewidth, height=.9in]{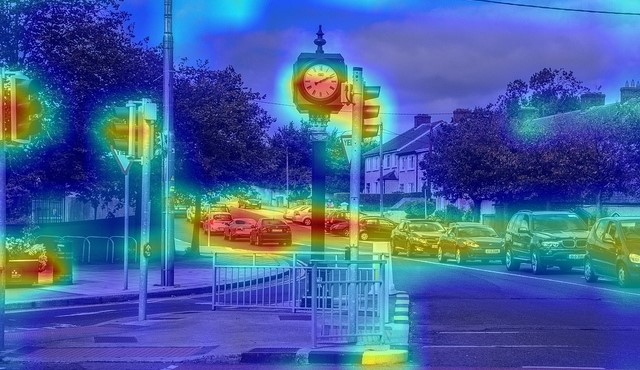}
   \end{subfigure}\\  

   \begin{subfigure}{0.24\textwidth}
     \includegraphics[width=\linewidth, height=.9in]{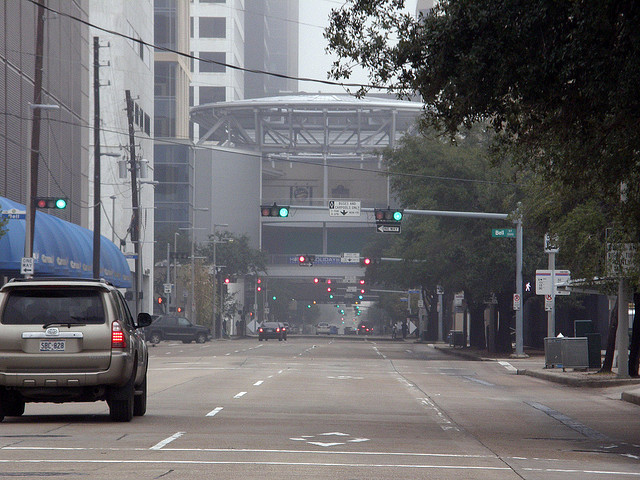}
     \caption{Original image} \label{fig:att_clean}   
   \end{subfigure}     
   \begin{subfigure}{0.24\textwidth}
     \includegraphics[width=\linewidth, height=.9in]{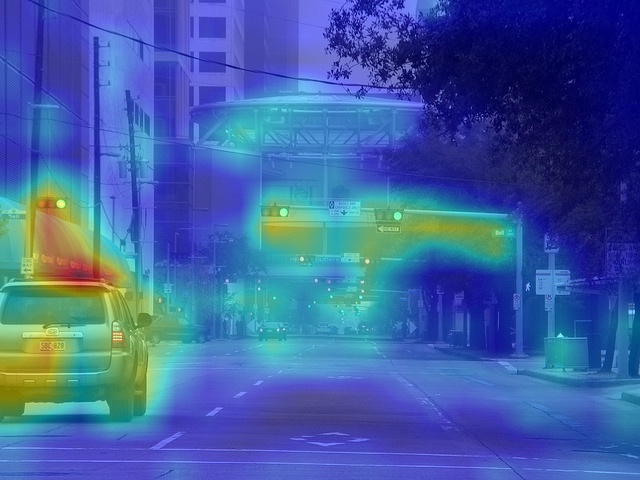}
     \caption{Baseline} \label{fig:att_orig_atss}
   \end{subfigure}
   \begin{subfigure}{0.24\textwidth}
     \includegraphics[width=\linewidth, height=.9in]{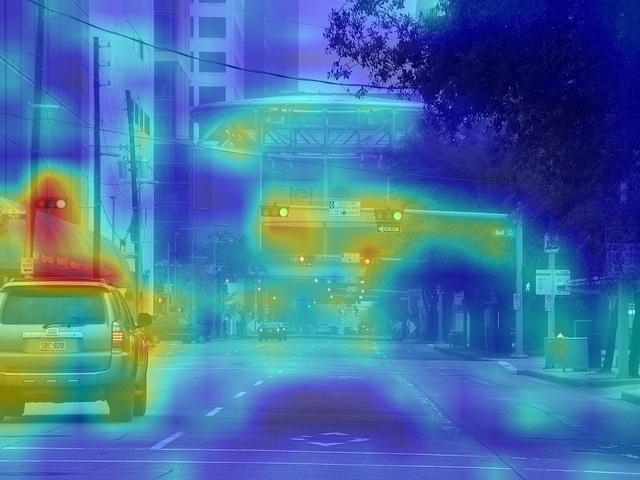}
     \caption{LDT} \label{fig:att_lda_atss}
   \end{subfigure}
   \begin{subfigure}{0.24\textwidth}
     \includegraphics[width=\linewidth, height=.9in]{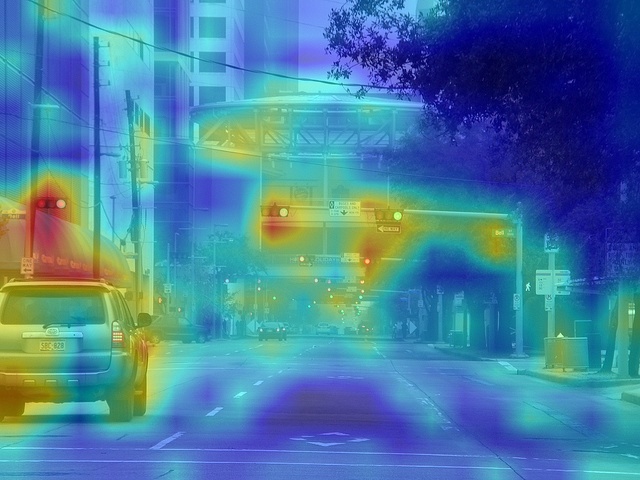}
     \caption{LDT-Pruned} \label{fig:att_pruned_atss}
   \end{subfigure}
\end{center}
\caption{
Attention maps from (b) the base model, (c) our model trained with LDT, and (d) our pruned model. Both (c) our LDT-trained model and (d) our pruned model pay more attention to relevant objects (red indicates the highest attention and blue the lowest).
Attention algorithm: Grad-CAM~\cite{Selvaraju2017}, base: YOLOF (top) and ATSS (bottom), pruning rate: 20\% (top), 10\% (bottom), data source: COCO.} \label{fig:attention}
\end{figure*}

Table~\ref{tab:COCO} presents the pruning results on COCO using the ATSS, GFL, and YOLOF base detectors. According to the results, our approach surpasses the four state-of-the-art methods across different pruning rates. The mAP improvements are up to 1.5\% for ATSS, 2.6\% for GFL, and 2.7\% for YOLOF. Furthermore, our pruned YOLOF model, with a 30.0\% reduction in FLOPs, increases mAP by 0.3\% compared to the original unpruned detector.
More importantly, as shown in Table~\ref{tab:COCO}, our approach, using even one iteration, demonstrates superior performance to the fine-grained GroupFisher method that retrains after each channel is pruned.
To maintain a comparable mAP to the unpruned base, the room for pruning on COCO is smaller than that on KITTI due to COCO's greater complexity.
\begin{table}[!htp]
  \centering
  \addtolength{\tabcolsep}{-1.5pt}
    \footnotesize 
    \begin{tabular}{l|cc|cc|cc}
        \toprule
          \multirow{2}{*}{Detector} & \multicolumn{2}{c|}{ATSS} & \multicolumn{2}{c|}{GFL} & \multicolumn{2}{c}{YOLOF}\\
        \cmidrule{2-3} \cmidrule{4-5} \cmidrule{6-7}
          & PR(\%) & mAP & PR(\%) & mAP & PR(\%) & mAP\\
        \midrule 
          Original & - & 39.4 & - & 40.2 & - & 37.5\\
        \midrule
          HRank
          & 15.0 & 38.8 & 15.1 & 38.5 & 20.0 & 37.9\\
          GroupFisher
          & 15.0 & 38.6 & 15.0 & 39.5 & 20.9 & 38.0\\   
          CHIP
          & 15.1 & 39.0 & 15.0 & 39.1 & 20.0 & 38.1\\
          PX & 15.1 & 38.7 & 15.0 & 39.4 & 20.5 & 37.9\\
          \rowcolor{lightgray!40}
          \textbf{Ours} 
          & 15.0 & \textbf{39.5} & 15.0 & \textbf{39.8} & 20.0 & \textbf{38.1}\\                
        \midrule
          HRank
          & 20.0 & 37.8 & 20.1 & 37.2 & 30.0 & 36.0\\
          GroupFisher
          & 20.0 & 38.5 & 20.0 & 39.3 & 29.0 & 37.5\\   
          CHIP
          & 20.0 & 37.7 & 20.0 & 38.3 & 30.0 & 36.9\\
          PX & 20.1 & 38.3 & 20.0 & 39.1 & 30.2 & 37.4\\
          \rowcolor{lightgray!40}
          \textbf{Ours} 
          & 20.0 & \textbf{39.2} & 20.0 & \textbf{39.8} & 30.0 & \textbf{37.8}\\         
        \midrule
          HRank
          & 25.0 & 36.5 & 25.0 & 36.1  & 40.0 & 34.4\\
          GroupFisher
          & 25.0 & 38.3 & 25.0 & 39.2 & 43.4 & 36.5\\   
          CHIP
          & 25.0 & 36.2 & 25.0 & 36.8 & 40.0 & 36.6\\
          PX & 25.0 & 37.9 & 25.0 & 38.6 & 40.1 & 35.8\\
          \rowcolor{lightgray!40}
          \textbf{Ours} 
          & 25.0 & \textbf{38.7} & 25.0 & \textbf{39.4} & 40.0 & \textbf{37.1}\\  
        \bottomrule
    \end{tabular}
    
    \caption{mAP performance of pruned ATSS, GFL, and YOLOF detectors on COCO. 
    PR represents the pruning rate, and we align FLOPs for fair comparisons.}
    \label{tab:COCO}
\end{table}

\subsection{Qualitative Results}

\subsubsection{Attention maps}

Fig.~\ref{fig:attention} shows example attention maps derived from three models: (b) the original base detector, (c) the detector trained with our LDT, and (d) our LDT-pruned detector. The attention maps demonstrate the influence of our LDT training and pruning. Both our LDT-trained detector (Fig.~\ref{fig:attention} c) and our LDT-pruned detector (Fig.~\ref{fig:attention} d) notably focus more on relevant objects compared to the original baseline model (Fig.~\ref{fig:attention} b). This is because our LDT training suppresses, while our LDT-based pruning removes, the effects of irrelevant and distracting dimensions. It is not trivial because the lack of focus to relevant objects often leads to missed detections (e.g., traffic lights).

\begin{figure*}[!htp]
\begin{center}

   \begin{subfigure}{0.19\textwidth}
     \includegraphics[width=\linewidth,  trim={7cm 1cm 20cm 2.8cm},clip]{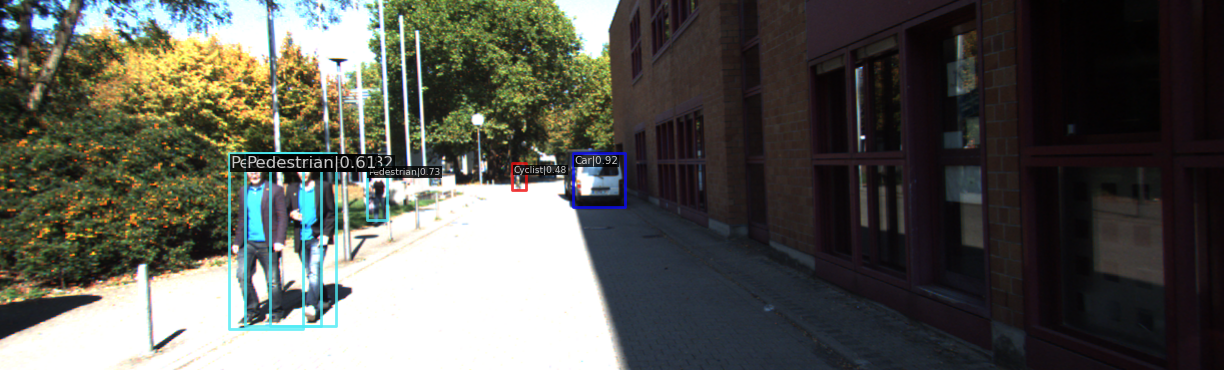}
   \end{subfigure}
   \begin{subfigure}{0.19\textwidth}
     \includegraphics[width=\linewidth,  trim={7cm 1cm 20cm 2.8cm},clip]{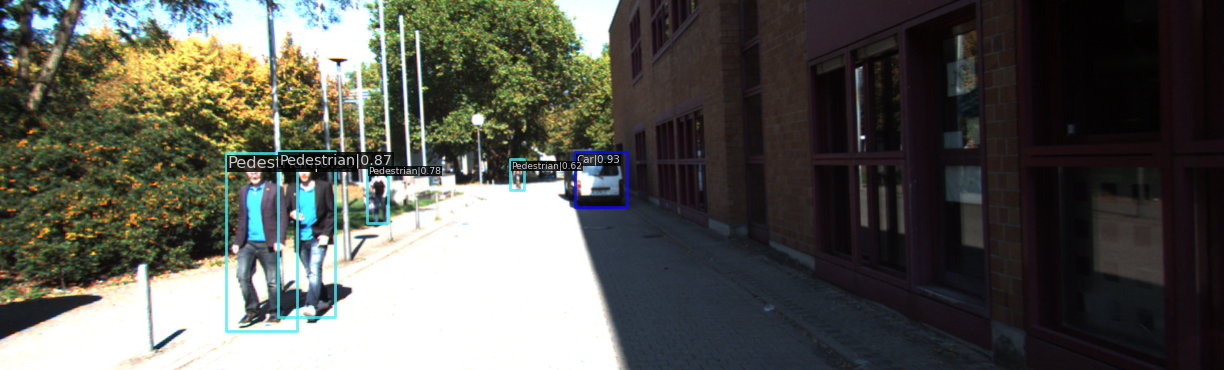}
   \end{subfigure}
   \begin{subfigure}{0.19\textwidth}
     \includegraphics[width=\linewidth,  trim={7cm 1cm 20cm 2.8cm},clip]{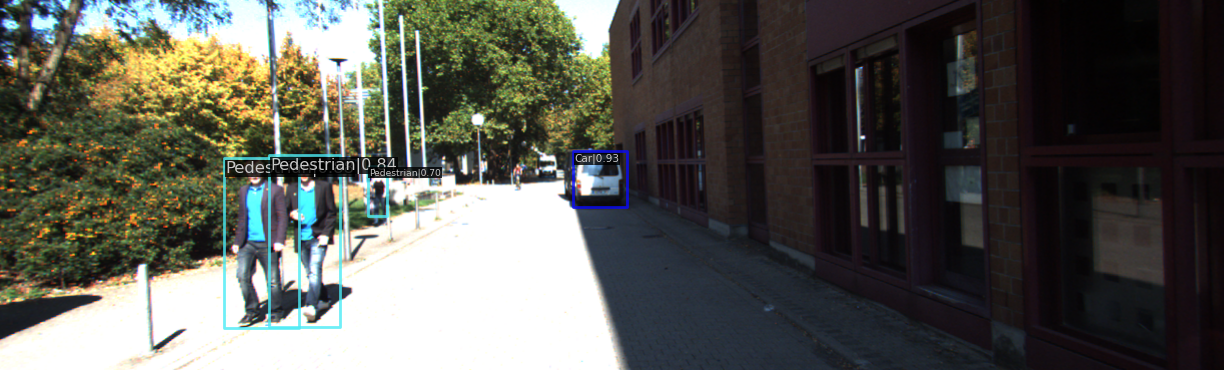}
   \end{subfigure}
   \begin{subfigure}{0.19\textwidth}
     \includegraphics[width=\linewidth,  trim={7cm 1cm 20cm 2.8cm},clip]{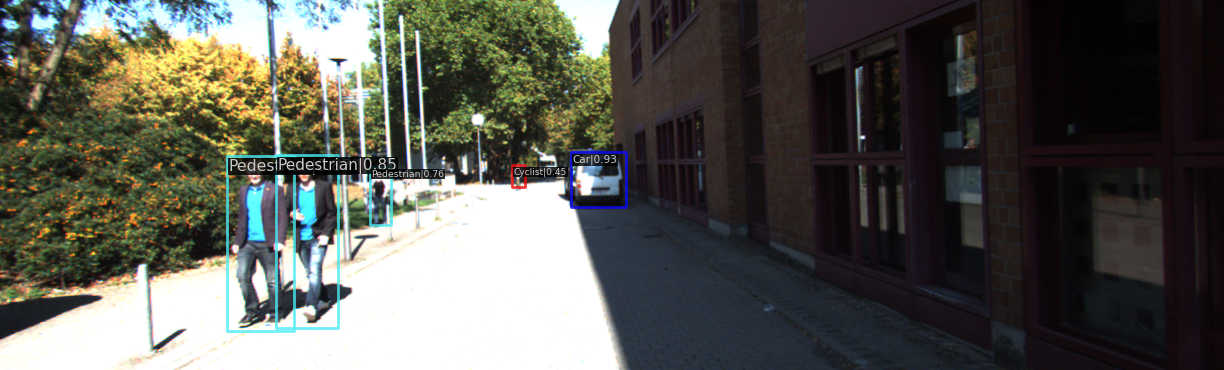}
   \end{subfigure}  
   \begin{subfigure}{0.19\textwidth}
     \includegraphics[width=\linewidth,  trim={5cm 0.8cm 15cm 1.9cm},clip]{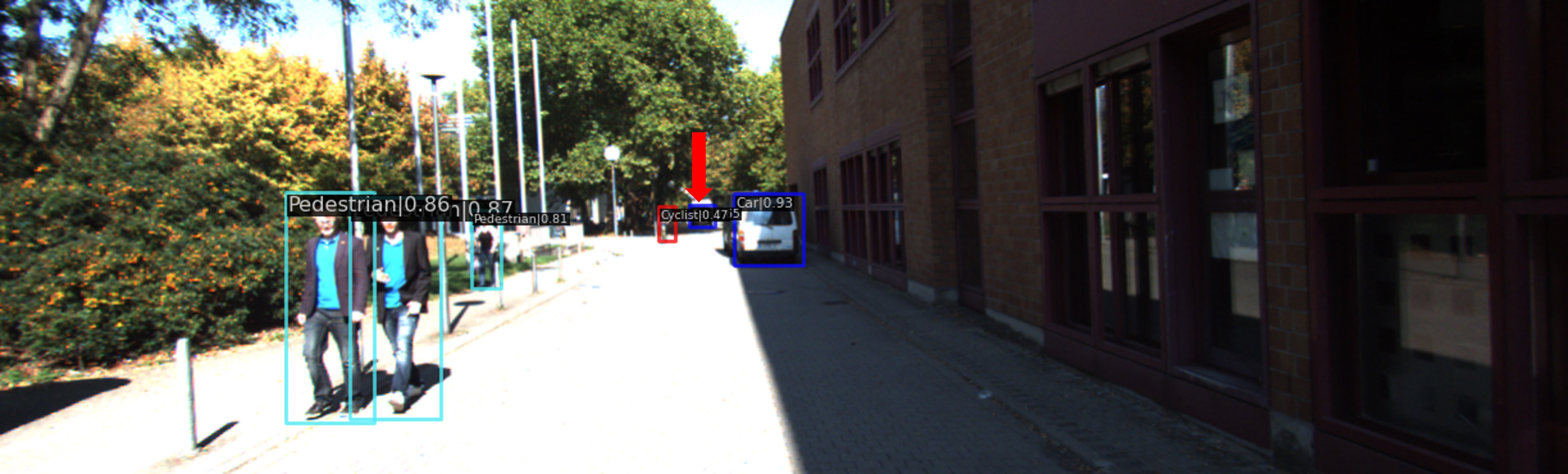}
   \end{subfigure}\\
  
   \begin{subfigure}{0.19\textwidth}
     \includegraphics[width=\linewidth,  trim={1cm 0cm 3cm 3.5cm},clip]{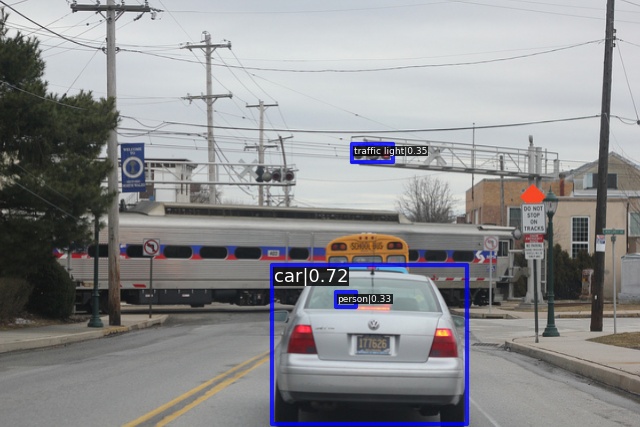}
     \caption{HRank
     } \label{fig:coco_ex_hrank}
   \end{subfigure}
   \begin{subfigure}{0.19\textwidth}
     \includegraphics[width=\linewidth, trim={1cm 0cm 3cm 3.5cm},clip]{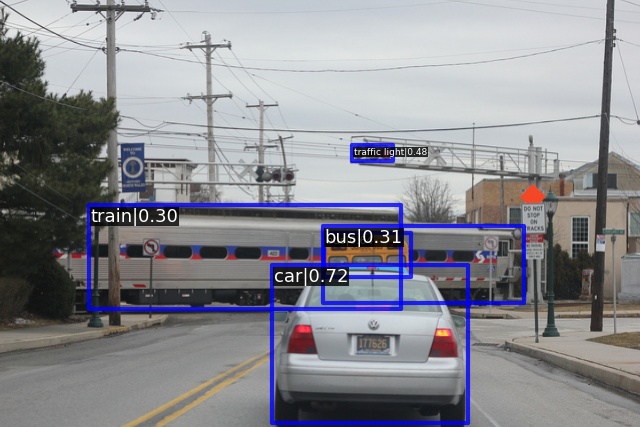}
     \caption{GroupFisher
     } \label{fig:coco_ex_gf}
   \end{subfigure}
   \begin{subfigure}{0.19\textwidth}
     \includegraphics[width=\linewidth, trim={1cm 0cm 3cm 3.5cm},clip]{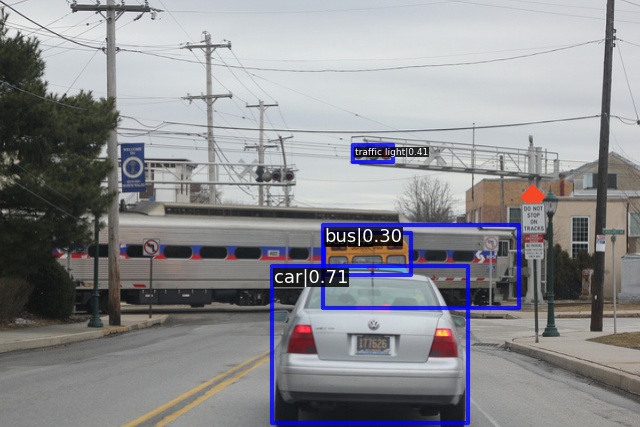}
     \caption{CHIP
     } \label{fig:coco_ex_chip}
   \end{subfigure}
   \begin{subfigure}{0.19\textwidth}
     \includegraphics[width=\linewidth, trim={1cm 0cm 3cm 3.5cm},clip]{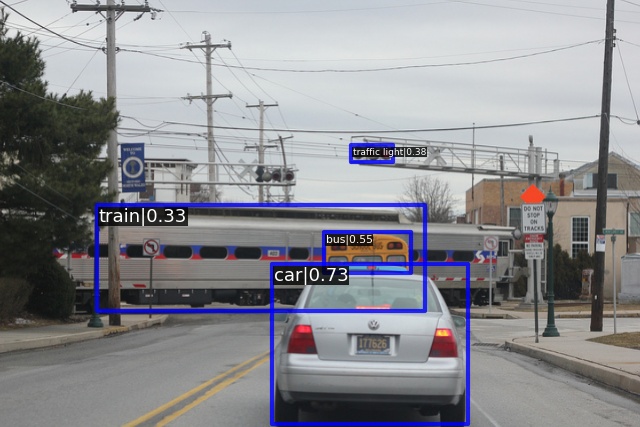}
     \caption{PX
     } \label{fig:coco_ex_px}
   \end{subfigure}
   \begin{subfigure}{0.19\textwidth}
     \includegraphics[width=\linewidth, trim={1cm 0cm 3cm 3.5cm},clip]{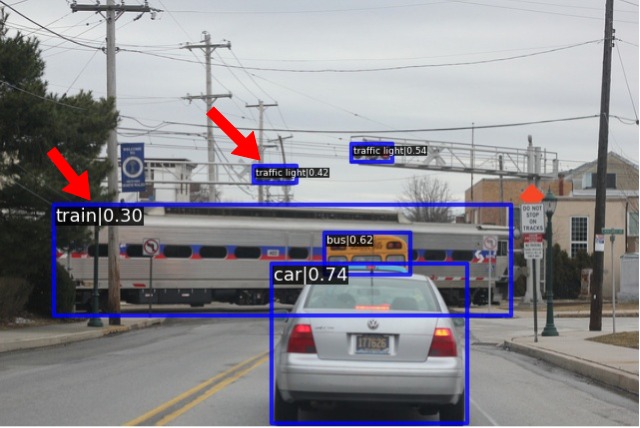}
     \caption{Our method} \label{fig:coco_ex_ours}
   \end{subfigure}
\end{center}
\caption{Qualitative results of (a) HRank~\cite{lin2020hrank}, (b) GroupFisher~\cite{liu2021group}, (c) CHIP~\cite{Sui2021}, (d) PX~\cite{iurada2024finding}, and (e) our approach. The red arrows indicate correct detections made by our pruned model but missed by competing models. The detection results in the top row are produced by a YOLOX-S model with 65\% of its parameters pruned on the KITTI dataset, while those in the bottom row are generated by an ATSS model with 18\% of its parameters pruned on the COCO dataset.} \label{fig:detection}
\end{figure*}
\subsubsection{Detection results}

In Fig.~\ref{fig:detection}, we qualitatively compare our pruning method's detection results with those of the four advanced competing approaches: (a) HRank, (b) GroupFisher, (c) CHIP, and (d) PX. The red arrows in (e) highlight our approach's superiority. This performance advantage stems from the minimal distractions in our compressed feature space, allowing our lightweight model to allocate its limited capacity to the most relevant aspects—an area that remains challenging for competitors.

\subsection{Ablation Studies}
\label{subsec:ablation}

\subsubsection{LDT}

In addition to better preparing the base detector for structured pruning, we evaluate LDT's impact on the base detector's performance. As shown in Table~\ref{tab:ablation_study_lda}, incorporating our LDT into visual detector training leads to improved performance across all detectors tested on both KITTI and COCO. In particular, the YOLOX-S and YOLOF models trained with our LDT outperform their originals by 1.6\% and 1.5\% mAP, respectively.
Notably, LDT adds only $\sim$10\% offline training overhead relative to its benefits.

\begin{table}[!htb]
\begin{subtable}[h]{.23\textwidth}
\footnotesize
\centering
\addtolength{\tabcolsep}{-5pt}
\begin{tabular}{c|cccc}
\toprule
Detector  & ATSS & GFL& YOLOX-S \\
\midrule
baseline & 89.7 & 89.7 & 90.7\\
\rowcolor{lightgray!40}
w/ LDT & \textbf{90.0} & \textbf{90.3} & \textbf{92.3}\\
\bottomrule
\end{tabular}
\caption{KITTI}
\label{tab:ldt_kitti}
\end{subtable}
\hspace{0.05in}
\begin{subtable}[h]{.22\textwidth}
\footnotesize
\centering
\addtolength{\tabcolsep}{-5pt}
\begin{tabular}{c|cccc}
\toprule
Detector  & ATSS & GFL& YOLOF \\
\midrule
baseline & 39.4 & 40.2 & 37.5\\
\rowcolor{lightgray!40}
w/ LDT & \textbf{40.2} & \textbf{40.5} & \textbf{39.0}\\
\bottomrule
\end{tabular}
\caption{COCO}
\label{tab:ldt_coco}
\end{subtable}
\caption{mAP Performance of deep detectors trained with and without our LDT on the (a) KITTI and (b) COCO datasets. Integrating our LDT into visual detector training enhances the performance of all tested detectors on both datasets.} 
\label{tab:ablation_study_lda}
\end{table}

\subsubsection{Utility source and location info}

We analyze the influence of the two factors on compressed model performance. After our location-aware discriminant training (LDT) is done, we compare the performance of channel utility originating from the neck ($LDT_{neck}$) with that derived from the final detection loss ($LDT_{det}$). According to Table~\ref{tab:ablation_study_merged}, channel importance derived from our LDT-compressed neck discriminants ($LDT_{neck}$) yields better mAP performance than using $LDT_{det}$. Moreover, incorporating localization information further enhances the performance. The pruned model, derived considering both localization and $LDT_{neck}$ information, enjoys the best performance and beats the unpruned base model by nearly 2\%.

\begin{table}[!htb]
\footnotesize
\centering
\addtolength{\tabcolsep}{-2pt}
\begin{tabular}{c|c|c|c|c}
\toprule
YOLOX-S & $LDT_{det}$ & $LDT_{neck}$ & $location$ & mAP\\
\midrule
Baseline  & - & - & - & 90.7\\

\midrule
\multirow{4}{*}{\shortstack{Pruned model \\ w/ LDT}}  & $\checkmark$ & & & 92.0\\
 &  $\checkmark$ & & $\checkmark$ &  92.3\\
 &  & $\checkmark$ &  & 92.2\\
 \rowcolor{lightgray!40}
 &  & $\checkmark$ & $\checkmark$ &  \textbf{92.6 (Ours)}\\
\bottomrule
\end{tabular}

\caption{Effect of utility source and location information on pruned detector performance. Our pruned model, using $LDT_{neck}$ and localization information, achieves the best performance. In $LDT_{neck}$, cross-layer utility originates from LDT-optimized neck discriminants, while $LDT_{det}$ uses utility from the final detection loss after LDT training. Base: YOLOX-S, dataset: KITTI, pruning rate: 20\%.} 
\label{tab:ablation_study_merged}
\end{table}

\subsubsection{Data efficiency and utility stability}

\begin{figure}[!htp]
\begin{center}
   \begin{subfigure}{0.32\linewidth} 
     \includegraphics[width=\linewidth]{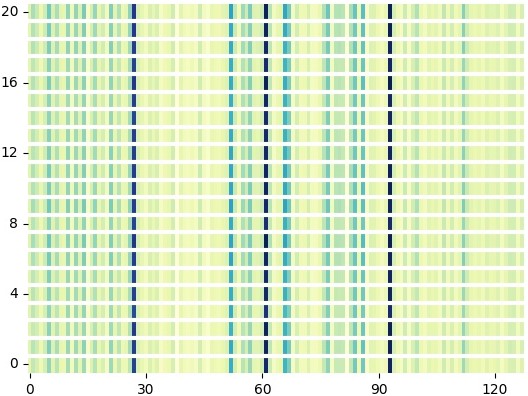}%
     \caption{stage2.1} \label{fig:s2}
   \end{subfigure}
   \begin{subfigure}{0.32\linewidth} 
     \includegraphics[width=\linewidth]{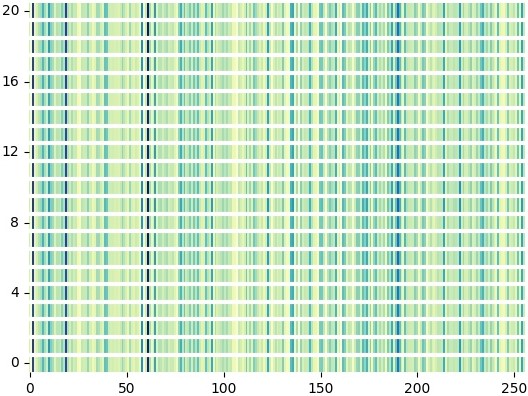}%
     \caption{stage3.1} \label{fig:s3}
   \end{subfigure}
   \begin{subfigure}{0.32\linewidth} 
     \includegraphics[width=\linewidth]{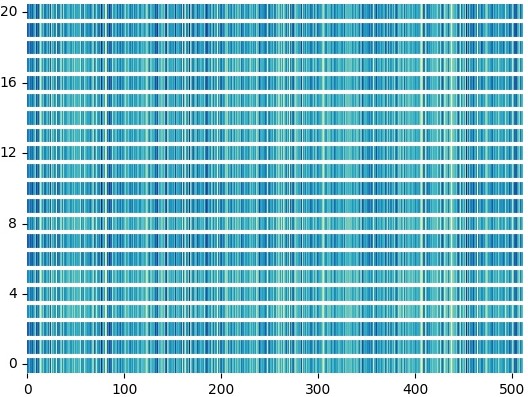}%
     \caption{stage4.2} \label{fig:s4}
   \end{subfigure}\\  
   
   \begin{subfigure}{0.32\linewidth} 
     \includegraphics[width=\linewidth]{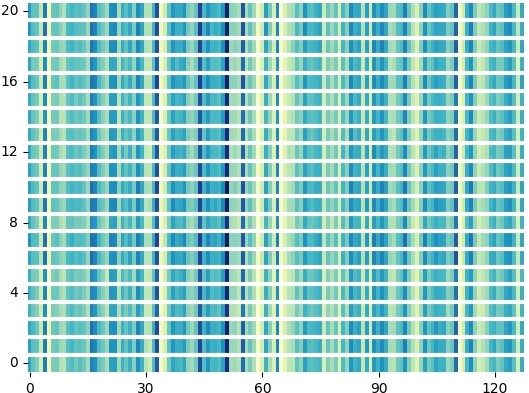}%
     \caption{top\_down\_bl.1} \label{fig:td_1}
   \end{subfigure}
   \begin{subfigure}{0.32\linewidth} 
     \includegraphics[width=\linewidth]{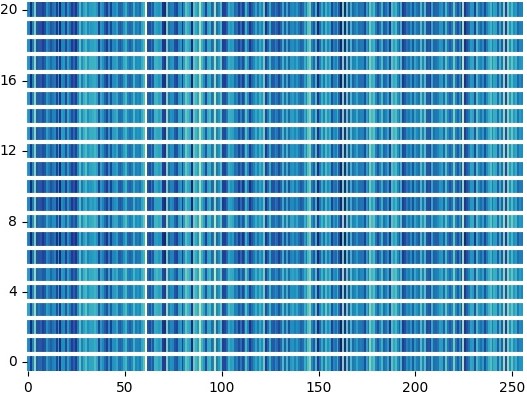}%
     \caption{bottom\_up\_bl.0} \label{fig:bu_0}
   \end{subfigure}
   \begin{subfigure}{0.32\linewidth} 
     \includegraphics[width=\linewidth]{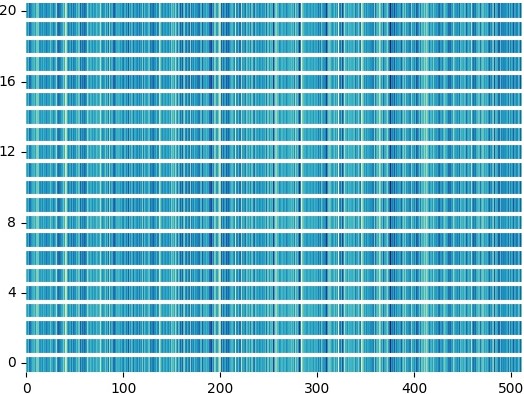}%
     \caption{bottom\_up\_bl.1} \label{fig:bu_1}
   \end{subfigure}
\end{center}

\caption{Channel importance of different stages'/blocks' final layers. The x-axis indicates channel indices, and the y-axis represents batch indices. Across different batches, the channels' average utility remains almost the same (indicated by color). Base: YOLOX-S, dataset: KITTI, batch size: 32.}
\label{fig:sensitivity}
\end{figure}

Our importance measure is data-driven. However, for large datasets, using the entire dataset to calculate channel importance is computationally expensive (and often unnecessary). Instead, we can utilize a subset of training images to achieve comparable performance. 
Through an example, Fig.~\ref{fig:sensitivity} illustrates that the average detection-related utility of each channel remains nearly constant across different batches of images. 
Since batch selection has minimal impact on a channel's utility, a subset of input images is usually sufficient to estimate its importance.

\subsection{Inference Latency Analysis}
In this section, we analyze how pruning-induced reductions in parameters and FLOPs translate into changes in inference latency. As shown in Table~\ref{tab:latency_kitti}, our pruned ATSS and GFL models achieve substantial computational savings, with FLOPs reduced by approximately 41\% to 49\% and parameters by around 70\%. These reductions lead to notable improvements in inference speed, reducing CPU latency by 36\% and GPU latency by nearly 30\%.
For YOLOX-S, pruning results in a 62.1\% decrease in FLOPs and a 74.8\% reduction in parameters, significantly improving inference efficiency. CPU latency decreases from 120.5 ms to 79.5 ms, yielding a speedup ratio of 1.52, while GPU latency drops from 18.1 ms to 15.1 ms, corresponding to a speedup ratio of 1.20. These improvements are achieved while maintaining comparable mAP performance; in fact, our pruning even enhances detector performance in the ATSS and GFL cases.

\begin{table}[h]
\footnotesize
\centering
\addtolength{\tabcolsep}{-2pt}
\begin{tabular}{cc|ccccc}
\toprule
\multicolumn{2}{c|}{\multirow{3}{*}{Detector}} & \multirow{2}{*}{mAP} & \multirow{2}{*}{FLOPs} & \multirow{2}{*}{Params} & CPU & GPU \\
& & & & & latency & latency \\
& & (\%) & (G) & (M) & (ms) & (ms)\\ 
 \midrule
\multirow{2}{*}{ATSS} & Base & 89.7& 219.56 & 32.12 &2865.2 & 48.9\\
& Pruned & 90.2 & 128.28 & 9.63 &1824.4 & 34.6\\
\midrule
\multirow{2}{*}{GFL} & Base & 89.7& 222.94& 32.26 &2911.5 & 48.8\\
&Pruned &90.4 &113.25 &9.68 &1901.6 &34.2 \\
\midrule
\multirow{2}{*}{YOLOX-S} & Base & 90.7 & 26.65 & 8.94 & 120.5 & 18.1\\
& Pruned & 89.6 &  10.10 
& 2.25 
& 79.5 & 15.1 \\
\bottomrule
\end{tabular}
\caption{Inference time of different base and pruned detectors using an Intel Xeon E5-2680 v4 CPU and a NVIDIA P100 GPU. Dataset: KITTI. mAP, GFLOPs, and MParams are included for reference.}
\label{tab:latency_kitti}
\end{table}

\section{Conclusion}
\label{sec:conclusion}
In this paper, we have proposed a proactive compression approach for deep visual detectors based on location-aware discriminant analysis. Through LDT training, detection-related discriminants are refined and aligned with a compressed set of latent channels, enabling and making more room for structured pruning. We trace the detection-related discriminating power throughout the network using derivative-based dependency analysis to guide the pruning process. Moreover, during deep discriminant analysis and detection utility tracing, we explicitly incorporate object location information, resulting in enhanced performance of our pruned detectors. Experiments on the KITTI and COCO datasets with various base detectors and state-of-the-art pruning methods demonstrate the superiority of our proposed approach. 
The plug-and-play nature of our method demonstrates its potential for widespread adoption in resource-constrained environments. Given the goal of edge deployment, this paper focuses on CNN-based detectors, owing to their relative efficiency (attributable to locality bias, analogous to low-power human visual perception). Future work will investigate extending the method to transformer-based detectors.

\section*{Acknowledgment}
This work was supported by the National Science Foundation (NSF) under Award Nos. 2153404 and 2412285. The PI, Tian, transitioned from BGSU to UAB during the course of the project. Computing resources were partially provided by the Ohio Supercomputer Center.
{
    \small
    \bibliographystyle{ieeenat_fullname}
    \bibliography{main}
}

\end{document}